\relax
\documentclass[letterpaper]{article} 
\usepackage{times}  
\usepackage{helvet} 
\usepackage{courier}  
\usepackage[hyphens]{url}  
\usepackage{graphicx} 
\usepackage{natbib}  
\usepackage{caption} 
\usepackage{tabulary}
\usepackage{natbib}
\usepackage{authblk}

\title{Codes, Patterns and Shapes of Contemporary Online Antisemitism and Conspiracy Narratives - an Annotation Guide and Labeled German-Language Dataset in the Context of COVID-19}

\author[1]{Elisabeth Steffen}
\author[1]{Helena Mihaljević}
\author[1]{Milena Pustet}
\author[1]{Nyco Bischoff}
\author[2]{María do Mar Castro Varela}
\author[2]{Yener Bayramoğlu}
\author[2]{Bahar Oghalai}
\affil[1]{Hochschule für Technik und Wirtschaft Berlin, Germany}
\affil[2]{Alice Salomon Hochschule Berlin, Germany}
\date{}                     
\begin{document}
\bibliographystyle{plainnat}

\maketitle

\begin{abstract}
Over the course of the COVID-19 pandemic, existing conspiracy theories were refreshed and new ones were created, often interwoven with antisemitic narratives, stereotypes and codes. The sheer volume of antisemitic and conspiracy theory content on the Internet makes data-driven algorithmic approaches essential for anti-discrimination organizations and researchers alike. However, the manifestation and dissemination of these two interrelated phenomena is still quite under-researched in scholarly empirical research of large text corpora. Algorithmic approaches for the detection and classification of specific contents usually require labeled datasets, annotated based on conceptually sound guidelines. While there is a growing number of  datasets for the more general phenomenon of hate speech, the development of corpora and annotation guidelines for antisemitic and conspiracy content is still in its infancy, especially for languages other than English. 
We contribute to closing this gap by developing an annotation guide for antisemitic and conspiracy theory online content in the context of the COVID-19 pandemic. We provide working definitions, including specific forms of antisemitism such as encoded and post-Holocaust antisemitism. We use these to annotate a German-language dataset consisting of ~3,700 Telegram messages sent between 03/2020 and 12/2021.
\end{abstract}

\section{Introduction}
Since its beginning, the COVID-19 pandemic is accompanied by an `infodemic', in the course of which vast amounts of misinformation, hate speech, rumors and conspiracy theories are being spread, in particular through social and online media \cite[19]{depoux_pandemic_2020,the_lancet_infectious_diseases_covid-19_2020}. Due to their oftentimes racist or antisemitic content, many of the conspiracy theories contribute to an increase in discrimination and even violence against the targeted groups 
\citep{bundesverband_rias_ev_antisemitismus_2020,gover_anti-asian_2020,meisner_sundenbocke_2021}. 

The unprecedented role of digital technologies and social media in the spread of conspiracy theories and hate speech present a key difference to previous pandemics: Respective narratives are being shared on video platforms, social networks, and messenger services, and with them racism, antisemitism, and calls for violence, which sometimes translate into violent attacks in the real world.\footnote{The assaults in Christchurch, Halle, or Hanau were impelled by racist and antisemitic conspiracy narratives disseminated via different online platforms, and the killers used online platforms to stage their killing in live streams \citep{musyla_christchurch_halle_hanau_2020}}

The sheer volume and the rapidly evolving online dissemination of antisemitism and conspiracy theory content make data-driven algorithmic approaches indispensable \citep{marcellino_detecting_2021}.
Researchers, anti-discrimination or fact checking organizations require technical support in identifying corresponding comments or articles on a large scale. Current publicly available services for automated detection of related phenomena such as toxic language, however, do not adequately cover antisemitism, in particular when it is communicated using codes and metaphors \citep{steffen_toxicity_2022}. 
In order to improve underlying machine learning models, comprehensive labeled data from online and social media are required. 

Existing datasets related to conspiracy theories or antisemitism are typically generated by filtering texts using explicit keywords such as `5G', `Bill Gates' or `jew*'. However, such approaches introduce a keyword bias to the generated corpora that makes it difficult to detect lesser-known or new conspiracy narratives or to identify intentionally obfuscated or coded terms, the latter being increasingly used e.g. to evade regulation by platform operators. Especially antisemitic content is often conveyed in an encoded way, using metaphors and codes that work without explicit reference to Jews or Israel \citep{zannettou_quantitative_2020,becker_decoding_2021-1}. In addition, newer forms of antisemitism are on the rise \citep{schmalenberger_tertiary_2022,schneider_querdenken-demo_2020} which are not sufficiently covered by standard working definitions, and difficult to discover through commonly used keywords.
This can also be observed in the context of the COVID-19 pandemic, when some anti-restriction protesters compare themselves with victims of the Shoah or equate the mandatory use of face masks with the obligation for Jewish citizens to wear the `Yellow Star' in Nazi-Germany \citep{schneider_querdenken-demo_2020}. 

In this paper, we draw on extensive research to develop an annotation guide for antisemitism and conspiracy theories in online content in the context of the COVID-19 pandemic. Regarding antisemitism, we focus on encoded forms of antisemitism and post-Holocaust antisemitism. We develop our annotation scheme as an interdisciplinary team to ensure a comprehensive conceptual approach. We provide real-world examples with our working definition to allow for its further development and its adaptation to other cultural, historical, and linguistic contexts and additional data sources. Furthermore, we use our working definition to annotate a German-language dataset \textit{TelCovACT} consisting of 3,663 Telegram messages posted between March 11, 2020 and December 19, 2021 and thus promote research in a less studied language. We chose Telegram because of its popularity among opponents of the government's measures to combat the coronavirus and the frequent spread of conspiracy theories and antisemitic statements \citep{european_commission_directorate_general_for_justice_and_consumers_rise_2021,winter_uberdosis_2021}. The dataset is made available to foster further research, especially on automated detection of antisemitic and conspiracy-theory 
content. 

\section{Related Work}
Our literature review focuses on studies in which large amounts of data have been collected and analyzed, typically in conjunction with annotation efforts, in order to provide  an overview of existing datasets and associated definitions and categorization schemes.
  
\paragraph{Conspiracy theories in social media}
Some recent works provide openly accessible annotated datasets and use them, often as part of challenges, to develop models for automated  classification \citep{alam_fighting_2021,golbeck_fake_2018}. In this context, conspiracy theories tend to be considered rather a subcategory and are often used synonymously with rumors or misinformation \citep{serrano_nlp-based_2020}, so that existing datasets and annotation schemes  are not based on a common theoretical foundation. In a systematic literature review of recent research on conspiracy theories, only around a third of considered works provided a definition of the term, thus analyzing “conspiracy theories online without explicitly defining the main object of their research” \citep{mahl_conspiracy_2022}.

A frequently applied approach is to address specific known conspiracy theories and gather data by searching for selected keywords  
\citep{marcellino_detecting_2021,memon_characterizing_2020,moffitt_hunting_2021,serrano_nlp-based_2020} without discussing the labeling process in detail \citep{gerts_thought_2021,pogorelov_fakenews_2020} or by resorting to examples in order to provide a definition \citep{moffitt_hunting_2021}. In most cases, a few thousand records are labeled manually; \citet{marcellino_detecting_2021}, however, use a much larger basis of 150,000 texts by refining the combination of search terms that refer to four well-known conspiracy theories but the process is not fully clear. 

Classification models trained on such keyword-based datasets yield moderate to high accuracy 
and typically employ language models such as BERT. Part of the research makes their datasets 
and codebooks 
openly available.

In addition to the rather pragmatic approaches, some works provide a solid theoretical foundation of the subject, discussing the relations among concepts such as conspiracy theories, rumors or misinformation \citep{kou_conspiracy_2017,samory_government_2018,wood_propagating_2018}. The different approaches to defining conspiracy theories turn out to share many common conceptual elements, in particular the assumption of a ``secret plot between powerful people or organizations'' \citep{mahl_conspiracy_2022} that work deliberately for their own sake and against the common good \citep{uscinski_why_2020}. Based on an extensive literature review of definitions and categorizations of conspiracy theories, \citet{samory_government_2018} deduce that the majority of relevant research ``relies on agents, actions, and goals as key elements in defining conspiracy theories or conspiracy beliefs,'' making paradigmatic examples dispensable. Along these lines, \citet{kou_conspiracy_2017} provide an operational definition of a conspiracy theory about public health crises containing the following three criteria: 1) the theory includes an explanation of the causality behind an event, 2) the explanation refers to primary actors (individuals or organizations, `the Other') whose actions are being kept secret from the public, 3) the actions have a malicious purpose, harming the greater good in favor of the actor’s own agenda. We incorporate these findings into our definition of conspiracy theories. 
\paragraph{Conspiracy theories and antisemitism}
We could not identify any publicly available datasets connecting conspiracy theories and antisemitism, and it seems that antisemitism has only recently attracted the attention of research on conspiracy theories in social media. This is supported by \citet{mahl_conspiracy_2022} who found hat only 2.1\% of the recent empirical studies addressing single conspiracy narratives focus on antisemitic narratives. 

Yet, at the same time, antisemitic stereotypes play an important role in current conspiracy theory discourses surrounding the COVID-19 pandemic. An alarming prevalence of antisemitism, both among religious conspiracies showing an age-old religious superstition and within deep state conspiracies, has been found on Twitter, together with a prevailing Nazi-Germany rhetoric in numerous German tweets debating coronavirus health measures \citep{media_diversity_institute_antisemitism_2021}. In an analysis of different social media platforms, \citet{cohen_antisemitism_2021} find that Jews are the second most targeted group in toxic posts. 

The pandemic has led to new antisemitic conspiracy theories \citep{cohen_antisemitism_2021}, while recycling old stereotypes. The narrative of ``Jews ruling international financial, political and media institutions'' is identified as most dominant antisemitic conspiracy theory element across different European countries and social media platforms \cite[9]{european_commission_directorate_general_for_justice_and_consumers_rise_2021}. Similar findings are supported by  an analysis of the YouTube presence of three leading British conspiracy theory spreaders with direct connections to the far right: the ``West as a whole is portrayed as dominated by a ruthless and bloodthirsty elite, whose members are often referred to using racially charged terms such as `Zionists', `Rothschilds', or `Rothschild Zionists''' \cite[97]{allington_antisemitic_2021}.

A recent comprehensive report examined the links between COVID-19 anti-vaccination conspiracy theories and antisemitism in Twitter and Facebook in seven European countries between March and August 2021 \citep{media_diversity_institute_antisemitism_2021}. Their key findings include that (1) anti-vaxxers typically perceive themselves as victims and resort to Holocaust comparisons and the self-labeling as `the new Jews'; (2) references to and variations of established antisemitic conspiracy theories such as `The Great Reset' and `New World Order' play a significant role; and (3) antisemitic codes such as `globalists' are frequently used throughout Europe. This showcases why an awareness of antisemitic codes, the structure of antisemitic argumentations, and specific forms such as post-Holocaust antisemitism is relevant for a classification of circulating antisemitic conspiracy content. 	 	 	 	

\paragraph{Antisemitic online content}

Almost all scientific studies known to us that tackle large-scale annotation of texts with respect to antisemitism utilize the working definition by the International Holocaust Remembrance Alliance (IHRA) as the main basis for their coding schemes \citep{becker_decoding_2021,chandra_subverting_2021,guhl_online_2020,jikeli_annotating_2019,jikeli_toward_2022,schwarz-friesel_judenhass_2019}.\footnote{\citet{chandra_subverting_2021} combine the IHRA definition with Brustein’s categorization of antisemitism into political, economic, religious, and racial antisemitism.} As shown by \citet{jikeli_annotating_2019}, using an English-language corpus containing the word (stems) `Jew*'
or `Israel', the IHRA definition is well suited in such a setting to generate a gold standard corpus for antisemitic content. However, many parts of the IHRA definition need further elaboration and refinement in order to serve as an annotation basis for automatic detection systems, as argued by \citet{jikeli_toward_2022} who extend their previous work to build an annotation scheme with many examples based on a close reading of the definition, with a clarification of ``grey zones'' and extensive literature on antisemitic stereotypes. Similarly, the project `Decoding Antisemitism' states to use the IHRA definition as a conceptual framework but extend it with further categories \citep{becker_decoding_2021-1} in order to annotate comments of major media outlets. 

We found only few studies that address algorithmic detection of antisemitic online content. We identified the paper by Warner and Hirschberg (\citeyear{warner-hirschberg-2012-detecting}) as the earliest work in this regard, who used a so-called template-based strategy to extract features from text and then trained an SVM, obtaining an F1 score of $\sim\! 0.6$ on a custom dataset consisting of 9,000 paragraphs. \citet{guhl_online_2020} used a commercial software to train a classifier  for online content from the imageboard 4chan. The model achieved an F1-score of $\sim \! 0.76$ on a small and rather specific dataset. \citet{ozalp_antisemitism_2020} trained a supervised machine learning model using 853 manually annotated tweets to detect online “antagonistic content related to Jewish identity” in tweets containing certain keywords by UK-based users. \citet{de_smedt_online_2021} created a machine learning based system for scoring English and German language texts regarding the level of antisemitic toxicity. The model is based on a self-developed lexicon consisting of over 2,000 relevant words and phrases containing Nazi-Germany rhetoric, dehumanizing adjectives and verbs inciting to violence, far-right terminology, alt-right neologisms, coded language, and revived conspiracy theories. In \citet{chandra_subverting_2021}, a multimodal deep learning classification model is trained on text and images, with an F1-score of 0.71 for Twitter and even 0.9 on Gab. 

Large-scale analyses of antisemitism based on models guided by annotated datasets, including all previously mentioned, use corpora created with keyword filters. These are often related to Jewishness or Judaism (e.g. `jew*', `hebrew') \citep{chandra_subverting_2021,jikeli_annotating_2019,jikeli_toward_2022,ozalp_antisemitism_2020}, or the state Israel, or reflect known and sometimes platform-specific antisemitic slurs (e.g. `kike', `ZioNazi', `(((jew))))' \citep{zannettou_quantitative_2020}, or are associated with antisemitic stereotypes and narratives (e.g. `happy merchant', `6 million') \citep{guhl_online_2020}. The authors partly reflect the limitation induced by restricting to such keywords. \citet{ozalp_antisemitism_2020}, for instance, underline that ``much antisemitic hate speech comes in the form of conspiracy theories (or allusions to such theories) and image-based hate speech—such as memes—that would not be captured by these keywords''. \citet{jikeli_toward_2022} justify such a restriction with the otherwise low percentage of positive examples in the annotated dataset that is not affordable under highly limited time budgets. It is, however, noteworthy that despite such a restriction on the content of the respective corpora, the annotation process is described by some as difficult \citep{jikeli_toward_2022,ozalp_antisemitism_2020}

The studies presented so far are limited to English-language content. German-language texts are covered by Monika-Schwarz Friesel's large-scale empirical study on antisemitic online content based on a large variety of text corpora and the project Decoding Antisemitism that analyzes comments to articles in mainstream media outlets in English, German and French \citep{becker_decoding_2021-1}. 
Except for \citet{chandra_subverting_2021,jikeli_toward_2022}, the manually annotated datasets are not announced as publicly available (on request). 

\section{Elaboration of adequate working definitions}
Antisemitism and conspiracy theories are inherently complex phenomena that can be difficult to annotate, especially when expressed as short texts in messenger services or social media platforms \citep{ozalp_antisemitism_2020}. Thus, careful elaboration of underlying theories is necessary for reliable annotation of datasets \citep{ross_measuring_2016}. 

\subsection{Antisemitism}

As our basis for a working definition of antisemitism, we turn to the working definition of the International Holocaust Remembrance Alliance (IHRA): 
\begin{quote}
Antisemitism is a certain perception of Jews, which may be expressed as hatred toward Jews. Rhetorical and physical manifestations of antisemitism are directed toward Jewish or non-Jewish individuals and/or their property, toward Jewish community institutions and religious facilities. \citep{international_holocaust_remembrance_alliance_working_2016}
\end{quote}

This definition has been recognized and implemented by numerous countries, cities, governmental and non-governmental institutions in various political and social fields. It has also shown to be a viable ground to manually annotate corpora that carry an explicit reference to Jews, Israel or antisemitic stereotypes \citep{jikeli_annotating_2019,jikeli_toward_2022}. 

As other researchers have pointed out, one of the strengths of the definition is that it covers most contemporary manifestations of antisemitism offering descriptive examples. However, the definition needs to be interpreted in context and its examples need to be concretized in order to use it as guidance for annotation \citep{jikeli_toward_2022}. For example, the definition does not (explicitly) address comparisons of current political measures of democratic governments to contain the pandemic with Nazi crimes against Jews.
Likewise, antisemitic narratives that address non-Jews are briefly mentioned in the definition but not further elaborated \citep{jikeli_annotating_2019}. Conspiracy theories regarding vaccinations are a particularly vivid example that combine these manifestations of antisemitism \citep{media_diversity_institute_antisemitism_2021}. We thus propose the following extensions and concretizations of the presented definition: 
\begin{enumerate}
	\item Post-Holocaust antisemitism (PHA) that refers to the instrumentalization of the victims of the Holocaust for a political agenda.
	\item Linguistic encodings of antisemitic statements that do not mention Jews or the State of Israel.	
\end{enumerate}

Our conceptualization aims at being both close enough to empirical data to be sufficiently context-sensitive for contemporary manifestations of antisemitic conspiracy narratives and at the same time abstract and generic enough to be adaptable to future research.

\paragraph{Post-Holocaust antisemitism}
The term post-Holocaust antisemitism (PHA) was coined by \citet{marin_post-holocaust_1980} to describe ``antisemitism without antisemites''. Corresponding narratives explicitly name Jews as part of argumentation strategies which instrumentalize the victims of the Holocaust for a political agenda and at the same time shift the perpetrator-victim coordinates by undertaking relativizing Holocaust comparisons. According to \citet{salzborn_verschworungsmythen_2021}, the instrumentalization is an essential component of this form of antisemitism. It fulfills a dual function: With regard to the past, it historically relativizes the Holocaust and infamously instrumentalizes the antisemitic policy of extermination. With regard to the present, it allows conspiracy theorists to demonize democratically legitimized rulings and measures by describing themselves as victims of a dictatorial state. 

In times of the COVID-19 pandemic, we encounter forms of PHA in comparisons or equations of the state measures to combat the pandemic with the National Socialist persecution of Jews. In Germany, some participants of demonstrations against restriction policies wear a yellow star with the imprint `ungeimpft' (unvaccinated) and thus symbolically compare themselves with Jews who under the National Socialist regime were forced to wear one. Leaflets and placards reading `Impfung macht frei' (vaccination sets you free), a reference to the slogan  `Arbeit macht frei' (work sets you free) at the entrance to Auschwitz and other Nazi concentration camps, were distributed or shown at protests \citep{belghaus_impfgegner_2020}. While this form of antisemitism first evolved in Germany (and Austria), where it has been analyzed as attempts of rejecting the guilt for the Shoah,  these narratives seem to have undergone a process of transnationalization in recent years, especially in the context of protests against COVID-19 measures \citep{media_diversity_institute_antisemitism_2021}.

\paragraph{Encoded antisemitism}
While the anonymity of online platforms on the one hand presents a fertile ground for explicit antisemitic hate speech, antisemitism is also often expressed via encoded, implicit manifestations. Findings from the ongoing research project Decoding Antisemitism which analyzes comments on German, French and British mainstream social media platforms indicate that “users use a variety of coded forms to communicate their antisemitic attributions” \cite[7]{becker_decoding_2021-1}, including semiotic markers such as icons or emoticons, abbreviations, word plays, allusions, and metaphors \cite[7]{becker_decoding_2021-1}.

Expressing antisemitic beliefs in an encoded, implicit form allows users to avoid social ostracism, the deletion of content from social media platforms, or even criminal consequences. 

Despite political and ideological differences, antisemitic discourses show a great uniformity and homogeneity regarding the stereotypes and codes used in them, highlighting their relevance for the transmission of antisemitism \citep{schwarz-friesel_judenhass_2019}. Encoded manifestations of antisemitism, via codes or metaphors, can thus also be assumed to play an important role for the online dissemination of antisemitism which is why it is of central importance to include them in annotation guidelines for antisemitic content. This also applies to manifestations of antisemitism in the context of COVID-19 conspiracy narratives: While on the one hand, the pandemic was framed by some conspiracy theorists as a smokescreen which was used by Zionists, the Rothschild family or George Soros to expand their power, other conspiracy narratives were even more popular: These narratives did not explicitly mention Jews as initiators of the pandemic and the resulting global crisis, but instead turned against Bill Gates, the `New World Order' or generally against `the' (economic or political) elites \citep{ajc_berlin_ramer_institute_antisemitische_2021,finkelstein_antisemitic_2020,european_commission_directorate_general_for_justice_and_consumers_rise_2021}. While the former narratives attribute special political and/or economic power to Jewish persons or groups, the latter operate encoded and get along completely without naming Jews.

\subsection{Conspiracy theories}
The term conspiracy theory was first coined by Popper, who argued that social sciences should not fall into the trap of providing simple explanations for unintended events, which he termed as `conspiracy theory of society' \cite[306]{popper_offene_2003}. According to Popper, unlike scientific explanations, conspiracy theories provide simple answers for complex social and political events.
Some conspiracy theories even refer to scientific studies as well as academic experts  to support their arguments. At the same time, a defining feature of conspiracy theories is their ``self-sealing quality'', meaning that they ``are extremely resistant to correction, certainly through direct denials of counterspeech by government officials.'' \citep{sunstein_conspiracy_2009}.

\paragraph{Working definition}
The concept of conspiracy theories is often used synonymously with similar forms of deceptive content such as disinformation (intentional dissemination of incorrect information) or misinformation (unintentional dissemination of incorrect information), rumors (unverified information), or fake news (fabricated news or a label used for delegitimizing news media) \citep{mahl_conspiracy_2022}. 

While conspiracy theories partially overlap with these concepts (e.g. a conspiracy theory might contain misinformation), they do have their own unique characteristics as attempts to create an alternative interpretation of events \citep{mahl_conspiracy_2022}: Conspiracy theories formulate the strong belief that a secret group of people, who have the evil goal of taking over an institution, a country or even the entire world, intentionally cause complex, and in most cases unsolved, events and phenomena \citep{butter_nichts_2018}. The exact intention (of a power-takeover) does not always have to be explicitly articulated; what is important, however, is the existence of a harmful intention and that the respective goal is of significant relevance to the public.\footnote{This does not include, in particular, theories without a specific harmful intent, such as the alleged existence of aliens covered up by governments.} A conspiracy theory can thus be considered an effort to explain some event or practice by reference to the machinations of powerful people, who have managed to conceal their role \citep{sunstein_conspiracy_2009}. Such a narrative is based on a simple dualism between good and evil which leaves no space for unintentional, unforeseeable things or mistakes to happen. Thus, a conspiracy theory needs actors (e.g. corrupt elites) who supposedly pursue a concrete malicious goal (e.g. control the population) using a strategy (e.g. by inserting a microchip via vaccinations) (see also \citet{samory_government_2018}).

The nature of social media and messenger services entails that more complex narratives are often incompletely rendered, especially when the counterpart can be assumed to be (partially) knowledgeable \citep{sadler_fragmented_2021,ernst_extreme_2017}. Accordingly, we believe that it is useful to annotate which of the components (actors, goal, strategy) actually appear in a given text. This will allow for post-annotation categorization of conspiracy theories in terms of completeness and fragmentation.

\paragraph{Conspiracy theories and the COVID-19 pandemic}
Studies show that particularly times of crises such as pandemics are prone to the emergence and spread of conspiracy theories \citep{heller_rumors_2015,kitta_vaccinations_2012,starbird_rumors_2014}. Since the threat posed by a disease is not directly tangible, pandemics often foster a range of conspiracy theories \citep{hepfer_verschworungstheorien_2015}. As an effect, the identified `culprits' can be named concretely and become tangible, which seemingly helps to structure an overwhelming situation. As was the case with other pandemics, the interpretations circulating on social networks lead to fatal mis- and disinformation about the origin and routes of infection or measures against the COVID-19 disease \citep{smallman_whom_2015}. In a similar way, a growing body of literature has observed how the outbreak of the pandemic not only led to a circulation of conspiracy theories but also how such theories led to the catalyzation or emergence of transnational movements such as QAnon and the so called Querdenken movement \citep{bodner_covid-19_2020}.

\paragraph{Compatibility with antisemitism}
There is a high degree of compatibility between antisemitism and conspiracy theories that is largely due to the strong structural ties between these two phenomena. \citet[49-50]{haury_antisemitismus_2019} elaborates the following fundamental principles characterizing modern antisemitism which are also central for our understanding of conspiracy theories: 
\begin{itemize}
	\item A specific form of personification, which attributes subjectless societal processes to the conscious, intentional, and malevolent actions of individuals; this inevitably induces the construction of an omnipotent enemy who has secretly taken over crucial points of control. 
	\item A Manichean worldview which radically divides the world into good and evil, a dualism based on an ontological construction of group identities. In this process, the enemies are represented as a foreign, external community with an immutable `nature' and characteristics. The usually nationally or ethnically constructed in-group is typically imagined as inherently good, naturally rooted, and free of internal conflicts or contradictions. 
	\item The enemy group is imagined as a corrosive and subversive threat for the in-group, potentially destroying its identity as well as its societal and political structures. Expulsion and extermination of the enemy group are seen as not only legitimate measures but as a last resort in face of the omnipotent, conspiratorial enemies who allegedly aim at destroying the collective. 
\end{itemize}
Conspiracy theories are thus  an ideal medium for the dissemination of antisemitic tropes, images and narratives. 

Looking at the narratives of protesters against COVID-19 countermeasures in Germany that reference conspiracy theories, it is possible to identify facets of all the outlined characteristics \citet{lelle_struktureller_2020}: 
Conspiracy theorists accusing individuals like Bill Gates, the (meanwhile former) chancellor Angela Merkel or virologist Christian Drosten of having a stake in and making profit out of the pandemic are examples of personification. Furthermore, the named persons are often accused of being part of a global conspiracy. The protesters perceive themselves as an `awakened' group which is fighting evil and spreading truth to disclose the lies of the global, malignant group of conspirators. Even though it must be noted that not all speakers at the demonstrations promote the idea of a nationally or ethnically defined community, most of them do emphasize the notion of community, unity and `naturalness'. As a consequence they contribute to a homogenization of the group of protestors. Furthermore, the frequent participation of Nazis and so-called `Reichsbürger' at the protests contributes to the spread of ethnic and nationalist ideologies in the movement. 

Finally, the increasing aggression, violent fantasies on posters and in chat groups, as well as the first violent attacks attributed to the spectrum of this movement point to its partial radicalization \citep{cemas_2021}. In some cases, this process is accompanied by a shift from structurally antisemitic attitudes to explicit and violent hatred of Jews \citep{rose_pandemic_hate_2021}.

\section{Corpus and annotation scheme}

We pre-selected all public channels identified by a research project to have a central role for mobilization against COVID-19 measures in the early phase of the pandemic \citep{forschungsinstitut_gesellschaftlicher_zusammenhalt_factsheet_2020} with more than 1,000 followers. In addition, we retrieved a Twitter dataset using keywords explicitly related to `Querdenken' from time periods around key demonstrations in 2020 and 2021 and identified all Telegram channels linked from it. The channels from both sources were then manually ranked for relevance to the research task using a random sample of 100 messages per channel. From the initial 215 channels, 133 were considered as particularly relevant.

We restricted to messages sent between March 11, 2020, the day COVID-19 was declared a pandemic by the WHO, and December 19, 2021. Very short messages or such with high similarity to other texts were excluded. Further details can be found in the datasheet documenting the dataset \citep{bischoff_etal_2022}.

\subsection{Annotation scheme}
Our annotation scheme is based on the working definitions of antisemitism and conspiracy theories presented above. 
As \textbf{two main categories}, we use the labels \textbf{`antisemitism'} and \textbf{`conspiracy theory'} to indicate that a message includes the respective content. For each main label, we provide \textbf{sub-labels} to annotate the \textbf{content} or \textbf{narrative structure} and the \textbf{stance} of a message if it was classified as antisemitism and/or conspiracy theory. 
 
 The provided \textbf{sub-labels for content} reflect our working definition of antisemitism, including \textbf{`encoded antisemitism', `post-Holocaust antisemitism', and `other forms of antisemitism'} to cover examples which would neither fit our definition of encoded nor post-Holocaust antisemitism. Annotators were encouraged to select only one content-related sub-label for antisemitism. Examples are provided in Table \ref{tab:1}.

For \textbf{conspiracy theory}, we used the narrative structure related sub-labels \textbf{`actor'}, \textbf{`strategy'}, \textbf{`goal'}, and \textbf{`reference'}, out of which the annotators could select all applicable labels. A good example for illustration is the first row in Table \ref{tab:1}, with `the satanic zionists' being the actor, `to kill billions of people' the goal, and `riots and fake pandemic' constituting the strategy. Regarding stance, we provided the sub-labels \textbf{`critical', `affirmative', `neutral or unclear'} for \textbf{antisemitism}, and \textbf{`authenticating', `directive', `rhetorical question', `disbelief'}, and \textbf{`neutral or uncertain'} for \textbf{conspiracy theory}, the latter based on an adapted form of the Rumor Interaction Analysis System (RIAS) scheme \citep{wood_propagating_2018}). 

We additionally labeled \textbf{`pandemic reference'} and \textbf{`GDR (German Democratic Republic) reference'} to indicate respective content. We furthermore used three technical labels: \textbf{`memorize task'} to mark a message for later consideration, e.g. because it was regarded as paradigmatic for a certain label; \textbf{`task unsuitable'} for messages annotators regarded generally unsuitable, e.g. because they contained sensitive information difficult to anonymize, were not German-language, or too short and thus incomprehensible; \textbf{`review required'} if annotators were uncertain how to classify a task and therefore wanted review by another annotator. Our detailed annotation guide is available at \citep{steffen_annotation_2022}.

\begin{table*}[]
\small
\begin{tabular}{p{0.07\textwidth} | p{0.85\textwidth}  } 
encoded & The long-term plan cooked up by the satanic zionists to kill billions of people is blowing up after the failure of their riots and fake pandemic. \\
\hline
PHA & The MASK now becomes the Yellow Star of the unvaccinated! 
A year ago it was a \#conspiracy theory that \#unvaccinated are marked separately. Today \#Lindner demands a \#mask obligation for all who are not \#vaccinated. Is the \#mask really becoming the new \#Jewish star? \\
\hline
other  & What has the Jew done to us? All ``vaccines'' are gene poison injections and come from Jewish corporations!  
\end{tabular}
\caption{Example texts communicating different forms of antisemitism.}
\label{tab:1}
\end{table*}

%


\subsection{Final dataset} 
Our dataset \textit{TelCovACT} consists of 3,663 records, approximately 14\% of which are labeled as antisemitic and 36\% as communicating conspiracy theories. At least one conspiracy-theory (antisemitism) related message was identified in 101 (85) of 133 channels. 
For almost all texts containing antisemitic content, the stance was classified as affirmative (94\%). Almost 60\% were labeled as encoded\footnote{Note that multiple labels were possible.}, making it the most frequent sub-form of antisemitism in the corpus. For conspiracy theories, belief was the the most frequent stance (95\%), followed by `authenticating' (24\%). The narratives most often included a strategy (72\%) and an actor (64\%). It should be noted that conspiracy theories were mostly communicated in a fragmented way, with only 26\% containing all of actor, strategy and goal, while 13\% communicated the respective content using a reference only, such as `\#QAnon'.

More than 72\% of all texts labeled as antisemitic also contained conspiracy theory content, while the majority of conspiracy theory messages (71\%) were not labeled as antisemitic. This does not mean that the majority of the conspiracy theories were not antisemitic;  our annotation scheme requires more than a mention of an antisemitic conspiracy theory to be labeled as such. Considering the level of fragmentation in the communication of conspiracy theoretical content, with 13\% communicating via a reference only, it is not  surprising that antisemitism is expressed in a minority of conspiracy theory messages. It is worth noting that the distribution of sub-forms of antisemitism differs significantly (chi-squared test with p-value $<0.01$) depending on the presence of conspiracy theory content: In the group of texts communicating both, the PHA label is given to less than 9\%, followed by `other' with 27\%, and encoded antisemitism being the leading label with a frequency of 72\%. However, if no conspiracy content was present, other forms of antisemitism are found most frequently (56\%), followed by PHA in 25\% of all cases, and only 23\% labeled as encoded. 

We computed chi-square test scores in order to find words that are most significant for the two categories. As expected, in texts communicating conspiracy theories we find in particular references to well-known theories such as New World Order, the Great Reset, deep state or `plandemie' (referring to a planned pandemic), actors such as Bill Gates, freemasons, Soros or Clinton, and words indicating strategies or goals such as lie, execute or dictatorship. The most significant words that are positively correlated with texts communicating antisemitism include references to Jewish identity such as jew or jewish as well as frequent codes such as illuminati, Soros, Rothschild, freemason or satanic.

\subsection{Data ethics and privacy}
Our approach to handle data ethics, privacy and protection follows best practices as documented in \citep{rivers_ethical_2014}. This includes exclusively collecting publicly available data and preventing data from being used to identify authors: Even though Telegram's Terms of Service states that user names and ids cannot be linked to a user's phone number as the only personal data collected by Telegram, we chose to additionally anonymize the dataset by replacing user names, user-ids as well as links to such by USER. Furthermore we decided only to provide our annotated dataset on personal request for research purposes approved for our ethical standards, thereby preventing any attempt of abuse.
We also note that the annotator team comprised nine individuals with diverse socio-demographic backgrounds, working in various  disciplines (five in political science or sociology and four in data science) with different levels of academic training. 

\section{Annotation process and evaluation}


The annotators reflected on their annotation experiences and discussed examples of conflicting annotations in a workshop to gain insights into factors which had affected their annotation decisions. 
During the workshop, almost all discussed conflicts could be resolved. After the joint workshop, 445\footnote{Of the 500 texts originally selected at random, 55 were marked as unsuitable and thus excluded from the evaluation.} texts were labeled by two annotators per message and used to compute the inter-annotator reliability. Our results indicate solid agreement among annotators, with Cohen’s kappa being 0.7 for conspiracy theory and 0.84 for antisemitism. 

The main insights from the workshop (and resulting modifications to the initial annotation scheme) are the following:

\paragraph{Positively biased corpus}
The annotated dataset was generated exclusively from channels known to spread conspiracy theories and antisemitic content, and all annotators were aware of this `positive bias'. This contextual knowledge influenced how some interpreted a message. 
Furthermore, the dataset mainly consisted of messages affirming antisemitism or conspiracy theories, making the selection of stance labels appear rather obsolete. However, for more heterogeneous data sources, the differentiation by stance is generally a useful additional information that can be utilized for training classifiers \citep{marcellino_detecting_2021}. 

\paragraph{Antisemitism}
Annotators expressed discomfort with classifying a text as `no antisemitism’, a label that was provided in our initial version of the annotation scheme, arguing that this could be interpreted as confirming a given text to be antisemitism-free. Furthermore they explained that sometimes a text itself could not be classified as antisemitic following the provided definition, but nonetheless would contain certain antisemitic undertones. For the final annotation scheme, we thus removed the choices `no’ and `uncertain’ for both main categories and instead added a label `review required' for cases requiring exchange with others.  

It was discussed how to classify texts that do not match our definition of antisemitism but include references to antisemitic conspiracy theories such as QAnon. We had only provided a `reference' label for the conspiracy theory category, but not for antisemitism. Annotators discussed whether these messages should be classified as `encoded antisemitism’; however, it was argued that the appearance of a single term or code is not sufficient, even when it is often used as an antisemitic code, making it difficult to introduce such a label for antisemitism.  


Annotators perceived the sub-label `other’ as potentially trivializing because it invokes the impression of being used for `secondary’ forms of antisemitism, while being too coarse-grained as it subsumes a variety of antisemitic content. Since our aim was to focus on post-Holocaust antisemitism and encoded antisemitism as previously underexposed manifestations of antisemitism in existing annotation tasks, we nonetheless consider the use of the sub-label `other’ as adequate. Depending on the research focus, however, a differentiation of it should be considered, e.g. using existing annotation schemes as in \citet{jikeli_toward_2022}. 

Other discussions evolved around messages mentioning Israeli politics. While the content itself mostly could not be classified as antisemitic, some annotators saw the mere focus on Israel in the context of our channel selection as a clear indication of antisemitic bias. While such consideration of focus and agenda-setting is common for qualitative approaches like critical discourse analysis, we doubt that it can be transferred to the training of classifiers that typically work on single-message level and have no knowledge of the `overall tendency' of a channel (in fact, finding channels communicating a certain amount of respective content is a plausible application scenario of a classifier). Another example for such a controversy around the consideration of context was a message which described COVID-19 prevention measures as systematic discrimination and fascist. For some annotators, the use of the term `Faschismus' in a German-language COVID 19-context indicated a clear relativization of the Shoah. They argued that in a German context, the term fascism is widely used synonymously with the German national socialist regime, and thus interpreted this message as a manifestation of post-Holocaust antisemitism. Other annotators doubted this interpretation, arguing that the term fascism potentially describes different kinds of phenomena.

\paragraph{Conspiracy}
The provided definition with its division into the elements actor, strategy, and goal was overall perceived as helpful and comparatively easy to apply. At the same time, annotators stated they sometimes found it difficult to clearly separate strategy and goal. Furthermore, if they could not identify a goal, they were more hesitant to label a text as conspiracy theory. Moreover, several messages were observed in which actors were only implicitly mentioned, e.g. as `they' or `our enemies', which is why we consider it an important feature of our annotation scheme to include implicit mentions.

Some texts were classified as conspiracy theory even though neither actor nor strategy nor a goal could be identified. This applies for example to texts describing the great majority of society ignorant of the conspiracy, e.g. by referring to them as `Schlafschafe’ (sheeples), or calling for an awakening of the masses. It was suggested to include the element `target’ or `victim’ of a conspiracy to our definition to include these kinds of texts. Other examples missing the defined triple were texts suggesting that `the truth’ was generally disguised. It was argued that both types of messages should be labeled as conspiracy theory.

Additionally, some annotators decided to apply the sub-label `reference’ if they interpreted a message as conspiracy theory but perceived it as too implicit and fragmented to apply the actual definition. In various cases, it became evident that background knowledge had influenced the decision, for example if a message contained links to platforms known to be disseminating conspiracy theory content. 

A lot of texts  turned out to contain fake news, dis- or misinformation. For the sake of feasibility, however, we had deliberately decided against providing respective labels, since this would require thorough fact checking, in some cases even scientific expert knowledge. Nevertheless, such text fragments were perceived as important discursive elements of conspiracy theories by some annotators who chose to label them as 'reference’  as a workaround. 
These decisions contributed to a partly inconsistent application of this sub-label.

%

\section{Discussion and future work}

The fraction of post-Holocaust antisemitism (PHA) in our dataset was lower than expected from qualitative analyses of Twitter data and public protests. 
We assume that the lack of regulation and content moderation on Telegram allows to uncover more directly one's antisemitic views, while PHA occurs more frequently in regulated contexts. 

The high proportion of encoded antisemitism, in particular in connection with conspiracy theory content, confirms that the antisemitic codes of a `global elite' in control of global political and economic processes can easily be adapted for the expression of belief in conspiracy theories. The large association of the two phenomena underlines the importance of approaching antisemitism in large online corpora without the restriction to keywords explicitly referring to Jews, Jewishness or Jewish collectives or institutions.

It is also worth noting that classification models trained on a dataset like this with high overall toxicity are more likely to actually learn aspects specific to antisemitism or conspiracy theory and not to related but differing phenomena such as offensive language or hate speech. 

With respect to the annotation process, we found that a major factor for different annotators' assessments was the handling of the context of a message or the consideration of related concepts such as misinformation. In this context, it is worth noting that some research explicitly includes contextual information such as images or external links into the annotation decision \citep{jikeli_toward_2022}, which is particularly helpful when labeling short texts.  
However, this places additional demands on the training of classification models. An alternative might be to treat threads or sequences of texts as entities instead of single messages. 

It also became clear that for annotators with a stronger affiliation to qualitative disciplines, it feels unfamiliar, not to say problematic, if they are asked to take a binary yes/no decision when interpreting a text. On the other hand, the group discussion showed that it is possible to reach a shared understanding and interpretation based on the predefined categories provided in our annotation guide in most of the cases; however, direct exchange between annotators needs to be assured in the labeling process. 

We find it important to make these differences and difficulties transparent, because we consider them relevant for other interdisciplinary research as well. After all, the divergences demonstrate the complexity of annotating human-written artifacts, a task which inevitably reduces complex social phenomena to a simplified classification. With this, it will hardly ever be possible to dissolve all conflicts emerging among annotators.
These conflicts could also be made productive to foster explicit and careful choices of how to resolve annotator disagreements: As \citet{gordon_jury_2022} have pointed out, the question whose voices are being heard when providing data for machine learning algorithms is often still left implicit and typically resolved by a majority vote. We think that especially for complex social phenomena, this process should gain more attention in future research – last but not least because power relations and discrimination affect people differently and are thus received with more or less sensitivity by them. 

\small
\bibliography{references.bib}

\end{document}